\documentclass[lettersize,journal]{IEEEtran}
\usepackage{amsmath,amsfonts}
\usepackage{algorithmic}
\usepackage{algorithm}
\usepackage{array}
\usepackage{hyperref}
\usepackage{subcaption}
\usepackage{textcomp}
\usepackage{stfloats}
\usepackage{url}
\usepackage{verbatim}
\usepackage{graphicx}
\usepackage{cite}
\usepackage{wrapfig}

\newtheorem{definition}{\textbf{Definition}}[section]

\usepackage{booktabs}
\usepackage[table,xcdraw]{xcolor}
\usepackage{multirow}
\usepackage{xspace}
\newcommand{\eg}{e.g.\xspace}
\usepackage{pifont}

\begin{document}


\title{Towards Robust Endogenous Reasoning: Unifying Drift Adaptation in Non-Stationary Tuning}


\author{ Xiaoyu~Yang, 
         En Yu,
         Wei Duan,
         Jie Lu~\IEEEmembership{IEEE Fellow}
\thanks{Xiaoyu Yang, En Yu, Wei Duan and Jie Lu are with the Australian Artificial Intelligence Institute (AAII), Faculty of Engineering and Information Technology, University of Technology Sydney, Australia}
\thanks{This work has been submitted to the IEEE for possible publication. Copyright may betransferred without notice, after which this version may no longer be accessible.}
}



\maketitle

\begin{abstract}

Reinforcement Fine-Tuning (RFT) has established itself as a critical paradigm for the alignment of Multi-modal Large Language Models (MLLMs) with complex human values and domain-specific requirements.
Nevertheless, current research primarily focuses on mitigating exogenous distribution shifts arising from data-centric factors, the non-stationarity inherent in the endogenous reasoning remains largely unexplored. 
In this work, a critical vulnerability is revealed within MLLMs: they are highly susceptible to endogenous reasoning drift, across both thinking and perception perspectives. It manifests as unpredictable distribution changes that emerge spontaneously during the autoregressive generation process, independent of external environmental perturbations.
To adapt it, we first theoretically define endogenous reasoning drift within the RFT of MLLMs as the multi-modal concept drift.
In this context, this paper proposes Counterfactual Preference Optimization ++ (CPO++), a comprehensive and autonomous framework adapted to the multi-modal concept drift.
It integrates counterfactual reasoning with domain knowledge to execute controlled perturbations across thinking and perception, employing preference optimization to disentangle spurious correlations.
Extensive empirical evaluations across two highly dynamic and safety-critical domains: medical diagnosis and autonomous driving. They demonstrate that the proposed framework achieves superior performance in reasoning coherence, decision-making precision, and inherent robustness against extreme interference. The methodology also exhibits exceptional zero-shot cross-domain generalization, providing a principled foundation for reliable multi-modal reasoning in safety-critical applications.

\end{abstract}




\section{Introduction}


\IEEEPARstart{R}{einforcement} Fine-Tuning (RFT) has established itself as a critical paradigm for the alignment of Multi-modal Large Language Models (MLLMs) with intricate human values and specialized domain requirements \cite{trungReFTReasoningReinforced2024,liuVisualRFTVisualReinforcement2025}. Within safety-critical sectors, such as healthcare and autonomous driving, the capacity for interpretable and generalizable reasoning highlights the significant advantages of RFT \cite{tan2025reason}. Nevertheless, MLLMs are frequently deployed in non-stationary environments that are characterized by unpredictable distributional shifts. This challenge is further intensified in multi-modal contexts, where the model navigates the complex coupling of visual and textual streams. 
Such heterogeneous interplay often triggers endogenous reasoning drift. It represents the spontaneous cognitive misalignment that arises from the internal stochasticity of the autoregressive generative process. Consequently, the reasoning trajectory of the model progressively unmoors from its perceptual foundation, a transition that occurs independently of exogenous environmental factors.
Ultimately, this internal instability leads to a cascading propagation of errors, which severely compromises the reliability of autonomous systems in high-stakes applications.


Prior works have investigated drift from the perspective of external environments. 
One prominent line of research focuses on generalization drift, addressing performance degradation on data beyond the training distribution, focusing on the external distribution shift encountered during domain adaptation \cite{chu2025sft, jin2025rl, wang2025reinforcement}. 
A parallel line of work addresses reward drift, examining the tendency of policies to exploit the inaccuracies of reward models, which leads to a systemic deviation from true human intent \cite{miao2025information, fu2025reward, liu2025rrm}. 
Finally, investigations into policy drift and entropy collapse scrutinize the training dynamics where the policy deviates excessively from the reference model or suffers from a collapse in distributional diversity \cite{yu2025dapo, wang2024beyond, xu2024is, ren2025learning, kumar2025llm}. 
Although these paradigms provide robust mechanisms for stabilizing RFT against external perturbations, they fundamentally treat drift as a phenomenon induced by exogenous environmental or data-centric factors. Consequently, the internal instability of the thinking process remains unexplored, leaving a critical gap in understanding the endogenous reasoning drift that arises spontaneously within the model.

\begin{figure}[htb]
    \centering
    \begin{subfigure}[t]{0.48\textwidth}
        \centering
        \includegraphics[width=0.8\textwidth]{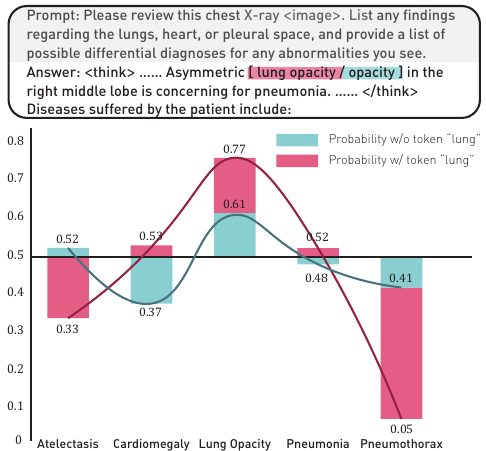}
        \caption{\textbf{Thinking Perspective.} Despite analogous occurrence probabilities and semantics of "lung opacity" (in red) and "opacity" (in blue) tokens within the chain-of-thought (CoT), slight perturbation of replacement induces significant adverse distributional drift in clinical conclusions, especially the opposite diagnosis of atelectasis, cardiomegaly and pneumonia.}
        \label{fig:reason-drift}
    \end{subfigure}
    \begin{subfigure}[t]{0.48\textwidth}
        \centering
        \includegraphics[width=0.8\textwidth]{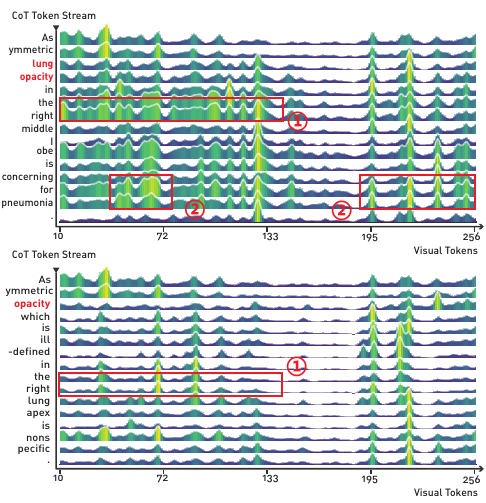}
        \caption{\textbf{Perception Perspective.} The attention scores over visual tokens are visualised during CoT decoding. When the term 'lung opacity' is subtly altered to 'opacity', we identify a substantial shift in the attention focus of visual areas within MLLMs, particularly in the highlighted regions.}
        \label{fig:visual-drift}
    \end{subfigure}    
    \caption{Endogenous Reasoning Drift in RFT.}
    \label{fig:toy-concept}
\end{figure}



Consequently, our motivation focuses on endogenous reasoning drift, which is revealed in the internal cognitive process inherent in the autoregressive generation of MLLMs. 
We posit that the autoregressive decoding paradigm can be formally characterized as a sequential token stream generation process. Within this framework, the thinking process is treated as a dynamic data stream that is susceptible to distribution shifts arising from evolving token probabilities and dynamic cross-modal attention allocation. We identify this internal instability as endogenous reasoning drift, a phenomenon that manifests through two interconnected dimensions: the thinking perspective and the perception perspective.


To illustrate the thinking perspective, we present a representative case study within clinical reasoning contexts as exhibited in Fig. \ref{fig:toy-concept}. 
For this investigation, Qwen2.5-VL \cite{qwen2.5-VL} is employed as the base model, with Direct Preference Optimization (DPO) \cite{rafailovDirectPreferenceOptimization2023} utilized to perform RFT on the MIMIC-CXR dataset \cite{johnson2019mimic}.  
As illustrated in Fig. \ref{fig:reason-drift}, we observe that the lexical selection between the terms 'lung opacity' and 'opacity' occurs stochastically during the generation of the thinking process. 
Crucially, even when lexical variants exhibit negligible probability and semantic differentiation during the thinking process, the resulting predictions for distinct pathologies can become diametrically opposed. This instability highlights a critical vulnerability where the reasoning trajectory of the model undergoes a systemic divergence, unmooring the final decision from its original logical premises.

We further extend the analysis to the visual modality to elucidate the perception perspective. Fig. \ref{fig:visual-drift} visualizes the cross-modal attention distribution during the generation of the thinking, plotting attention scores for 256 visual tokens encoded from the chest X-ray at each step of token generation. Brighter colors indicate higher attention activation, and the first ten visual tokens are masked to mitigate the attention sink phenomenon \cite{xiao2024efficient}. As presented in Fig. \ref{fig:visual-drift}, when the term lung opacity is subtly altered to opacity, a significant drift in visual attention occurs, causing the attention distribution of the MLLM to undergo a systemic collapse. 
Specifically, during the prediction of the localization phrase in the right, highlighted in \ding{172} of Fig. \ref{fig:visual-drift}, the robust attention activation originally across relevant pulmonary regions degrades into sparse, isolated spikes. 
Furthermore, during the inference of pneumonia, highlighted in \ding{173} of Fig. \ref{fig:visual-drift}, the attention on critical pathological regions completely vanishes. Such evidence suggests that endogenous reasoning drift in non-stationary RFT originates not only from cumulative linguistic biases but from a fundamental attentional shift, where the reasoning trajectory of the model loses its grounding in visual evidence.

In essence, endogenous reasoning drift represents an intrinsic instability within the cognitive process of MLLMs. It is defined by unpredictable distribution changes that emerge spontaneously during the autoregressive generation process, independent of external environmental perturbations. 
Ultimately, this internal drift creates a critical gap between the reasoning trajectory and the objective evidence, highlighting the imperative for a robust mechanism to stabilize the internal distributional characteristics of the model.



Our preliminary work Counterfactual Preference Optimization (CPO) \cite{yang2025walking}, published on NeurIPS 2025, established a link between the thinking instability of RFT and the theory of concept drift. By leveraging the principles of counterfactual reasoning \cite{pearl2000models,pearl2014interpretation}, CPO employs an auxiliary LLM to generate a hierarchical concept graph to synthesize perturbed CoT trajectories that remain linguistically coherent yet lead to contradictory conclusions. Through a dual-preference optimization objective, this mechanism compels the model to disentangle beneficial distribution adaptation from detrimental concept drift, thereby ensuring stable alignment within non-stationary environments.


Despite the advancements achieved by CPO \cite{yang2025walking}, the scope of that framework remains primarily confined to the thinking perspective of endogenous reasoning. By focusing exclusively on the linguistic and logical consistency of the reasoning trajectory through textual counterfactuals, CPO overlooks the perception perspective, particularly the stability of cross-modal grounding. In the context of MLLMs, reasoning is intrinsically coupled with visual perception, as exhibited in Fig.\ref{fig:visual-drift}. Therefore, addressing instability solely within the linguistic domain is insufficient to ensure total reliability. Without a mechanism to synchronize the stability of the thinking process with perceptual alignment, the model remains vulnerable to the systemic collapse of visual attention, which prevents the attainment of comprehensive robustness in non-stationary environments.

To overcome the limitations of existing methods, we propose Counterfactual Preference Optimization ++ (CPO++), a comprehensive adaptive framework designed to unify multi-modal concept drift adaptation within non-stationary environments. Primarily, to resolve the structural asymmetry found in the previous approach, the methodology of this work extends counterfactual disentanglement from the isolated domain of the thinking perspective to the perception perspective. 
Through the introduction of a novel visual-reasoning consistency protocol, the framework explicitly penalizes internal visual drift, thereby strictly anchoring the textual reasoning trajectory to authentic visual evidence. 
Furthermore, to eliminate dependency on static, predefined schemas and to enhance cross-domain applicability, a self-evolutionary mechanism is integrated into the architecture. This advancement empowers the foundational model to autonomously synthesize highly aligned preference thinking paths and precisely decoupled counterfactual trajectories. 
Guided by the embedding of domain knowledge via a hierarchical knowledge graph, this process operates without the requirement for external expert supervision.

In summary, our paper mainly makes the following contributions:

\begin{itemize}

    \item Theoretically, we formally define endogenous reasoning drift within the RFT of MLLMs as the multi-modal concept drift, which is characterized by unpredictable distribution changes that emerge spontaneously during the cognitive process.

    \item CPO++ framework is proposed to execute the counterfactual decoupling across both thinking and perception perspectives, thereby systematically eliminating confounding biases induced by endogenous reasoning drift. 


    \item Extensive empirical evaluations are conducted across the safety-critical domains of medical diagnosis and autonomous driving. It validates our method achieves superior performance in reasoning coherence, decision precision, inherent robustness, and zero-shot cross-domain generalization.

\end{itemize}



\section{Related Works}
\label{appendix:relatedwork}

\subsection{Reinforced Fine-tuning}


Besides, the landscape of alignment methodologies continues to diversify through innovative paradigms: ReST \cite{gulcehre2023reinforcedselftrainingrestlanguage} employs iterative self-generation of policy-derived samples to refine LLMs via offline reinforcement learning, while DPO \cite{rafailovDirectPreferenceOptimization2023} fundamentally reformulates alignment as direct preference optimization through implicit reward modeling. Concurrent developments span Rejection Sampling Fine-Tuning's \cite{yuan2024scaling} curation of validated reasoning trajectories for supervised augmentation, and ReFT's \cite{trungReFTReasoningReinforced2024} phased optimization combining SFT initialization with PPO-driven exploration of automated reasoning path generation. Building upon these foundations, Visual-RFT \cite{liuVisualRFTVisualReinforcement2025} extends GRPO-based strategies to multimodal contexts, enhancing visual-language alignment under data scarcity, whereas B-STaR \cite{zeng2025bstar} introduces dynamic configuration adaptation for self-teaching systems through principled exploration-exploitation balancing. Pushing the boundaries of evaluation rigor, Qwen-Math-PRM \cite{zhang2025lessons} synergizes Monte Carlo estimation with LLM-as-judge consensus filtering while pioneering a hierarchical assessment framework integrating stepwise and holistic performance metrics.
Moreover, SCALAR \cite{yang2024enhancing,young2025fewer} performs visual grounding unsupervised via reinforced learning under the open-world environment. Besides, APO \cite{yang2025learningallconceptalignment} leverages reinforcement learning to align the knowledge in multiple teacher models for distillation.

The integration of reinforcement learning (RL) into the post-training alignment of LLMs has undergone a remarkable evolution since the seminal work on Reinforcement Learning from Human Feedback (RLHF) \cite{christiano2017deep, ouyang2022training}. A critical component of this paradigm is Proximal Policy Optimization (PPO) \cite{schulman2017proximal}, which has established itself as the standard for policy optimization by ensuring training stability through the use of clipped probability ratios. While early implementations demonstrated the efficacy of modeling human preferences, the prohibitive costs of manual annotation have catalyzed a transition toward automated reward generation. This shift is illustrated by the approach of Bai et al. \cite{bai2022constitutional}, which utilizes sparse natural language feedback as proxy signals. More recently, the progressive framework of DeepSeek \cite{guo2025deepseek} established baseline performance through pure RL before introducing the R1 variant. This model achieves improved generalization through the cyclic alternation between supervised fine-tuning and the Group Relative Policy Optimization (GRPO) protocol \cite{shao2024deepseekmath}, exemplifying the progression of the field toward self-contained alignment systems.

The landscape of alignment methodologies continues to diversify through several innovative paradigms. Reinforcement Self-Training (ReST) \cite{gulcehre2023reinforcedselftrainingrestlanguage} employs the iterative self-generation of policy-derived samples to refine LLMs via offline RL. Simultaneously, Direct Preference Optimization (DPO) \cite{rafailovDirectPreferenceOptimization2023} fundamentally reformulates alignment as a direct optimization problem through implicit reward modeling, thereby removing the requirement for an explicit reward model. Concurrent developments include the curation of validated reasoning trajectories for supervised augmentation in Rejection Sampling Fine-Tuning \cite{yuan2024scaling}, and the phased optimization of ReFT \cite{trungReFTReasoningReinforced2024}, which combines SFT initialization with the exploration of automated reasoning paths. Building upon these foundations, Visual-RFT \cite{liuVisualRFTVisualReinforcement2025} extends strategies based on GRPO to multi-modal contexts to enhance visual-language alignment under conditions of data scarcity. In a similar vein, B-STaR \cite{zeng2025bstar} introduces dynamic configuration adaptation for self-teaching systems through the balancing of exploration and exploitation. To enhance evaluation rigor, Qwen-Math-PRM \cite{zhang2025lessons} synergizes Monte Carlo estimation with consensus filtering while pioneering a hierarchical assessment framework. Finally, SCALAR~\cite{yang2024enhancing} performs unsupervised visual grounding via reinforced learning in open-world environments.

\subsection{Multi-modal Reasoning}

Recent advances in Long Chain-of-Thought (Long CoT) reasoning \cite{li2025system,chen2025towards,cai2025unleashing,lu2026querycounselstructuredreasoning} have significantly enhanced the capacity of LLMs to perform multi-step reasoning and self-correction. By incorporating self-reflection strategies, these models can dynamically diagnose and revise intermediate reasoning traces, thereby mitigating certain types of reasoning inconsistencies during inference. In contrast, the approach of this work introduces a proactive intervention at the training stage through counterfactual sample generation, which explicitly shapes the causal representations and decision boundaries of the model. This enables the model to internalize more consistent causal reasoning patterns from the outset, rather than relying on post-hoc correction during inference. Hence, this methodology and Long CoT-based self-reflection strategies are viewed as complementary: while Long CoT enhances the reliability of reasoning at runtime, the proposed approach strengthens causal robustness during the learning phase. It is anticipated that future work could explore the integration of both directions—leveraging the reflective inference mechanisms of Long CoT together with counterfactual training—to further advance consistent causal reasoning in reasoning-oriented LLMs.

In parallel, multi-modal Chain-of-Thought (MM-CoT) reasoning \cite{zhang2023multimodal,chen2024m,wang2025multimodal,zheng2023ddcot,caisteering} has emerged as an important paradigm for enabling LLMs to align and reason across heterogeneous modalities such as vision and language. For instance, M3CoT~\cite{chen2024m} and Ddcot~\cite{zheng2023ddcot} introduce multi-domain, multi-step reasoning frameworks that emphasize structured cross-modal inference, while the survey in~\cite{wang2025multimodal} provides a comprehensive taxonomy of MM-CoT paradigms and benchmarks. Compared with these inference-time reasoning approaches, the proposed method introduces a proactive causal intervention during training through counterfactual sample generation, which aims to enhance causal robustness within multi-modal reasoning processes. This approach is complementary to both Long CoT and MM-CoT paradigms—while existing frameworks improve the reasoning trajectory during inference, this work focuses on stabilizing the causal representations during model optimization. This synergy represents a promising direction for future research.

Besides, several studies \cite{lu2025newborn,lu2026choosing} have integrated causal inference to enhance the interpretability and reliability of reasoning within MLLMs. For example, Li \textit{et al.} \cite{li2025revealing} explore the potential of LLMs in revealing complex multi-modal causality. To address the issue of model hallucinations, Zhou \textit{et al.} \cite{zhou2025mitigating} propose the deciphering of attention causality to mitigate hallucinations induced by modality priors. Furthermore, Causal-CoG \cite{Zhao_2024_CVPR} adopts a causal-effect perspective to optimize context generation, thereby enhancing the reasoning capabilities of multi-modal models.

\begin{figure*}[htbp]
    \centering
    \includegraphics[width=0.98\textwidth]{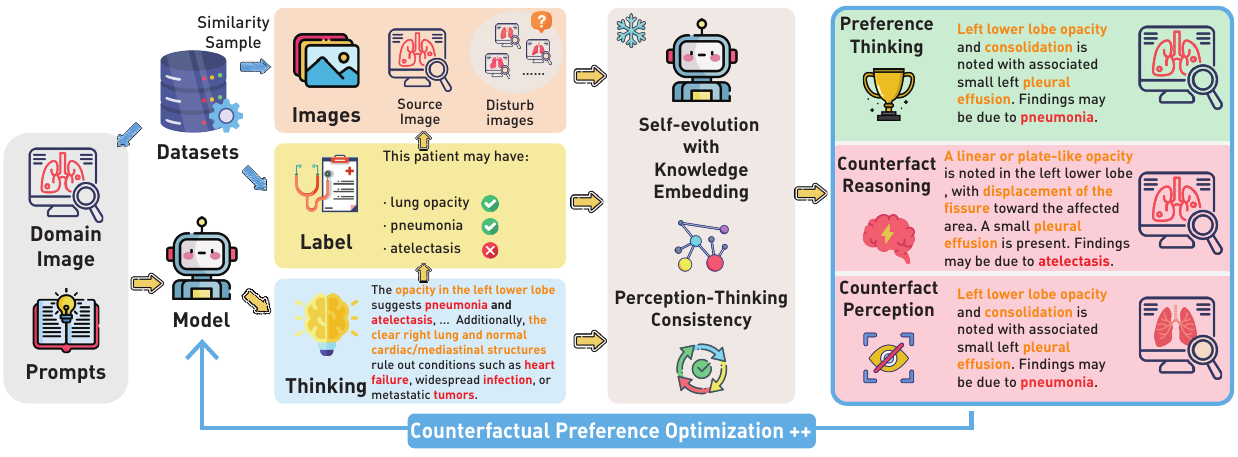}
    \caption{\textbf{The proposed Counterfactual Preference Optimization ++ (CPO++) framework.} 
    To mitigate endogenous reasoning drift, the methodology theoretically characterizes it as multi-modal concept drift, and incorporates counterfactual inference to disentangle spurious correlations from genuine causal logic within the original outputs. 
    By leveraging hierarchical domain knowledge and perception-thinking consistency protocol, the framework generates highly aligned preference thinking paths, along with precisely decoupled counterfactual reasoning and perception trajectories. 
    These constructed samples are subsequently utilized to drive the preference optimization of the model, thereby simultaneously adapting the endogenous reasoning drift across both thinking and perception perspectives.}
    \label{fig:workflow}
\end{figure*}

\subsection{Concept Drift}

Concept drift, fundamentally defined as unpredictable changes in underlying data distributions \cite{krawczyk2017ensemble, luLearningConceptDrift2019}. As the field evolved, the focus has shifted from traditional statistical learners to deep learning architectures, especially the MLLMs. Yuan et al. \cite{yuanRecentAdvancesConcept2022a} provide a systematic review of this transition, emphasizing the necessity of adapting deep neural networks to non-stationary environments where static assumptions no longer hold.
Besides, Mi \cite{miConceptNeuralNetwork2025} proposed dynamic stream learning architectures based on time-delay regret. Additionally, drift-aware mixture of experts systems have been introduced to manage the dynamic shifts inherent in large-scale data ingestion \cite{yu2026drift,yu2025generalized}.

The synergy between concept drift theory and reinforcement learning is increasingly vital for maintaining agent stability in non-stationary settings. Recent studies have demonstrated the efficacy of automated drift handling in cloud infrastructures \cite{shayestehAutomatedConceptDrift2022b} and the mitigation of adversarial shifts in malware detection \cite{mcfaddenDRMDDeepReinforcement2025}. Most relevant to this work, Yang et al. \cite{yang2025adapting,yangTdistributedSphericalFeature2023} established a precedent for adapting MLLMs to concept drift from the pre-training stage onwards, highlighting the persistent nature of distribution shifts in generative models.
Moreover, ResilientCL \cite{yangCausalInformedContrastiveLearning2025,yangMaskedImageContrastive2024} proposes the causal interventional contrastive objective to mitigate the concept drift within the momentum network of contrastive pre-training paradigm.

\section{Methodology}

\subsection{Overview}


To address the challenge of endogenous reasoning drift, we first establish a formal theoretical characterization of this phenomenon as multi-modal concept drift in  Section \ref{sec:3.3}. Within this theoretical framework, we incorporate counterfactual inference as a principled mechanism to disentangle spurious correlations from genuine causal logic, as exhibited in Section \ref{sec:3.4}. 
Combining the causes with a hierarchical domain knowledge, the proposed methodology synthesizes counterfactual perception-thinking trajectories that are significant domain-specific semantic relevant, in Section \ref{sec:3.5}. 
Ultimately, the Counterfactual Preference Optimization ++ (CPO++) drive the autonomous evolution of the model, thereby facilitating robust alignment and continuous self-improvement, as presented in Section \ref{sec:3.6}.

\subsection{Multi-Modal Concept Drift within MLLMs}
\label{sec:3.3}


In this section, we theoretically define the phenomenon of endogenous reasoning drift as multi-modal concept drift, characterized by unpredictable distribution changes that occur spontaneously during the autoregressive sequential decoding process of MLLMs. 
The formalization extends traditional concept drift to the domain of multi-modal, encompassing both thinking and perception perspectives.
It illustrates how the semantic alignment between perception evidence and thinking trajectories undergoes a progressive decoupling that introduces stochastic biases throughout the cognitive process.

Operating through recursive on-policy sampling, the MLLM $\pi$ autoregressively generates the token at position $j$ in the chain-of-thought. This process is explicitly conditioned on the multi-modal context, comprising the visual input image $v$, the textual prompt $l$, and the partial token sequence $t_{<j}$ of the CoT trajectory:
\begin{equation}
    t_{j} \sim \pi(\cdot | v, l, t_{<j})
\end{equation}
To capture the dynamics of cross-modality interactions, we formally extend the concept drift definition to the multi-modal reasoning process as follows:

\begin{definition}
    \label{def:multimodal_drift}
    \textit{
    The MLLM's autoregressive generation manifests as a multi-modal thinking stream $S_{0,i} = \{s_{0},...,s_{i}\}$. Each cognitive state $s_{j}=(t_{<j}, z_{j} | v)$ encapsulates the token sequence generated so far $t_{<j}$ and the latent predicted distribution $z_{j}$ of the final results, strictly conditioned on the perception evidence $v$. Consequently, at position $i$, the stream $S_{0,i}$ adheres to a visual-conditioned distribution $F_{0,i}(x, z | v)$. The multi-modal concept drift within the reasoning process is thus formalized as a distributional shift in the conditional joint probability:
    \begin{equation}
        \label{eq:dis}
        \exists i: P_{i}(t, z | v) \neq P_{i+1}(t, z | v)
    \end{equation}
    where the joint probability $P_{i}(t, z | v)$ can be further decomposed to explicitly reveal the perception-thinking alignment:
    \begin{equation}
        P_{i}(t, z | v) = \underbrace{P_{i}(t | v)}_{\text{Perception Grounding}} \times \underbrace{P_{i}(z | t, v)}_{\text{Thinking Consistency}}
    \end{equation}
    }
\end{definition}

Consequently, the multi-modal concept drift framework highlights that drift arises when the thinking trajectory $t$ progressively decouples from the perception evidence $v$ where $P_{i}(t | v)$ shifts, or when the mapping from thinking to prediction $z$ becomes unstable wherein $P_{i}(z | t, v)$ biases. Our proposed framework provides a unified theoretical lens to characterize the unpredictable dynamics of CoT reasoning.

To adapt the multi-modal concept drift characterized in Definition \ref{def:multimodal_drift}, specifically the progressive decoupling of thinking trajectories from perception evidence, it is imperative to enforce an alignment mechanism. 
Thus, the optimization objective is formulated to minimize the divergence between the model's evolving distribution and the optimal perception-grounded thinking trajectory. 
Formally, this adaptive reinforced custom-tuning objective is defined as:
\begin{equation}
\label{eq.goal}
    \min_{\pi^{(i)}, \dots, \pi^{(i+\tau)}} \sum_{k=i}^{i+\tau} \mathcal{L} \left( \prod_{j=1}^{L} \pi^{(k)}(t^{(k)}_{j} | v, l, t^{(k)}_{<j}) , y^{(k)} \right)
\end{equation}
where $\prod_{j=1}^{L} \pi^{(k)}(t^{(k)}_{j} | v, l, t^{(k)}_{<j})$ represents the joint probability of the generated CoT sequence strictly conditioned on the perception evidence $v$ and textual prompt $l$. 
Here, $y^{(k)}$ denotes the ground-truth perception-faithful trajectory, and $L$ signifies the maximum token length. The policy $\pi^{(k)}$ represents the state of the MLLM at the cognitive status $k$. By minimizing the cumulative loss metric $\mathcal{L}$ over the sliding time window $[i, i+\tau]$, the model is explicitly driven to rectify the probabilistic shifts in $P(t,z|v)$, thereby maintaining robust thinking trajectories despite the underlying non-stationarity.

\subsection{Disentangling Drift with Counterfactual Causes}
\label{sec:3.4}


The optimization objective formalized Eq.\ref{eq.goal} necessitates disentanglement of two competing goals: advantageous policy-induced domain adaptation and versus concept drift arising from suboptimal policy execution, which are both sampled from policy $\pi$ within the time period $[i, i+\tau]$. However, it is challenging to determine the optimal preferred CoT and explicitly judge which strategies will cause unpredictable changes in tokens.

\begin{figure}[htbp]
    \centering
    \includegraphics[width=0.4\textwidth]{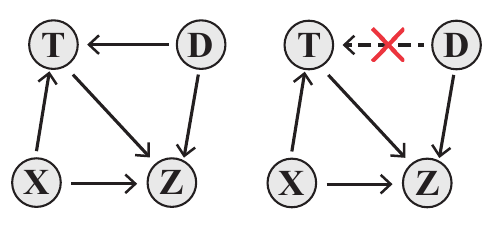}
    \caption{\textbf{Structural Causal Graph.}  \textbf{X}: Inputs, \textbf{Z}: Prediction Results, \textbf{T}: Chain-of-Thought, and \textbf{D}: Latent Concept Drift within Non-Stationary CoT.}
    \label{fig:causal}
\end{figure}

Fortunately, counterfactual causes provide an explicit manner to decouple these two competing goals. We construct a structural causal graph \cite{pearl1995causal,pearl2016causal} to formalize the causal relationship among elements as discussed in Section \ref{sec:3.3}, including inputs ($X$) consisting of image $v$ and prompt $l$, prediction result ($Z$), Chain-of-Thought ($T$) and the concept drift within CoT ($D$) as illustrated in Fig.~\ref{fig:causal}, where $A\rightarrow B$ denotes that $A$ is the direct cause of $B$. The causal graph of $\{X,Z,T,D\}$ presents the following causal connections:


$(X,D)\rightarrow T$: This link denotes the chain-of-thought $T$ derived from the inputs $X$ through policy $\pi$ is under the impact of latent concept drift $D$.

$(X,T,D)\rightarrow Z$: This link presents that, apart from the regular reasoning pathway of $(X, T) \rightarrow Z$, the prediction is also impacted by the concept drift $D$ through the pathway of $D\rightarrow T\rightarrow Z$.

In the constructed structural causal model,  nodes $D$ and $T$ are formally characterized as the confounder and mediator~\cite{pearl2022direct}, respectively. The confounding variable  $D$
induces bias through the backdoor path  $T\leftarrow D\rightarrow Z$~\cite{pearl1995causal}, simultaneously influencing both the mediator $T$ and the outcome variable $Z$. This interference systematically distorts the estimation of CoT reasoning's causal effect on model predictions, particularly under non-stationary adaptation scenarios. 
The resulting spurious correlations enter the mediation pathway $X\rightarrow T \rightarrow Z$ through the confounded mediator $T$, ultimately compromising the stability of customized model tuning.

Building upon the above analysis grounded in cause, we formally decouple the concept drift dynamics in Chain-of-Thought reasoning of Eq.~\eqref{eq.goal} grounded in the cause. By constructing interventional distributions through $\textbf{do}$ operations, $P(Z| \textbf{do}(T=t), D=d)$, we quantify a latent causal effect of Chain-of-Thought reasoning on model predictions:
\begin{equation}
\label{Eq.4}
    \psi = \mathbb{E}[Z_{T\leftarrow t, D\leftarrow d} - Z_{T\leftarrow t', D\leftarrow d}]
\end{equation}
where the potential outcome $T\leftarrow t$  represents the counterfactual scenario when forcibly maintaining the mediator $T$ at value $t$, while preserving the confounder state 
$d$. It corresponds to a controlled direct effect of the mediator $T$ on the outcome $Z$ under a fixed concept drift state $D=d$.
This formulation explicitly isolates the causal effect of Chain-of-Thought $T$ on prediction $Z$, from backdoor concept drift propagation $T\leftarrow D \rightarrow Z$.

\subsection{Self-Counterfact with Domain Knowledge}
\label{sec:3.5}

Having operationalized the causal disentanglement in Section~\ref{sec:3.4}, a critical challenge remains in practical implementation: how to acquire high-quality, causally aligned counterfactual instances without heavy reliance on external supervision. To address this, we propose a self-evolutionary framework that leverages the MLLM's intrinsic capabilities to generate, perturb, and verify reasoning chains through a structured knowledge graph and the perception-thinking consistency.

\subsubsection{Hierarchical Domain Knowledge Graph}


To ensure that the generated endogenous reasoning trajectories adhere to domain causal logic rather than spurious correlations, we first construct a hierarchical concept graph $\mathcal{G}_{K}$ that codifies domain-specific knowledge structures. It systematically organizes concepts into three universal semantic dimensions applicable across generalized tasks, such as medical diagnosis \cite{yang2024segmentation,yang2022local,chen2026tc} or autonomous driving:

\begin{figure}[htbp]
    \centering
    \includegraphics[width=1\linewidth]{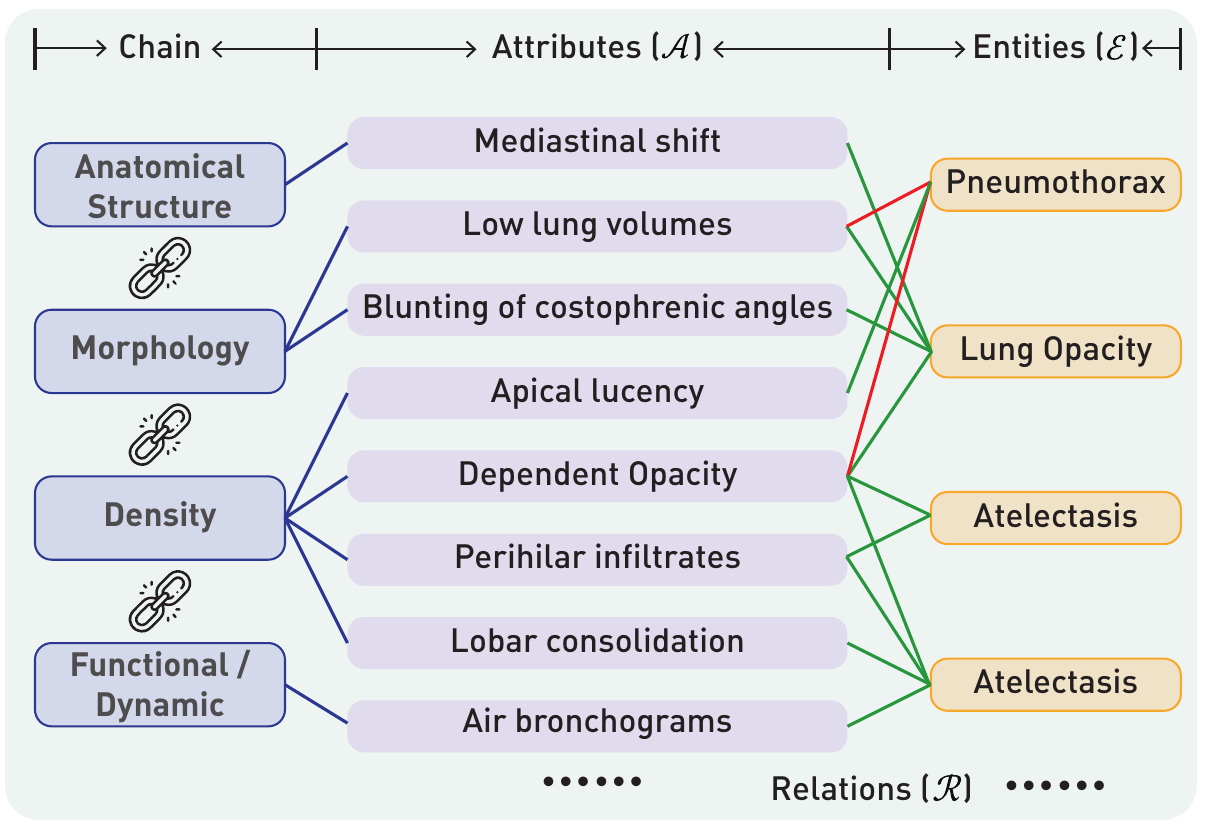}
    \caption{\textbf{Example Case of Hierarchical Domain Knowledge Graph in Medical Domain.} 
    To disentangle detrimental drift, we introduce the graph that generates plausible counterfactual CoTs through controlled attribute perturbations.
    Green lines represent attributes that are positively associated with the disease, while the red denotes that they are exclusive. 
    }
    \label{fig:know-graph}
\end{figure}

\begin{itemize}
    \item \textbf{Entities ($\mathcal{E}$):} The core objects of interest (e.g., \textit{Pneumonia} in radiology, or \textit{Car} in autonomous driving).
    \item \textbf{Attributes ($\mathcal{A}$):} The observable perception features defining entities (e.g., \textit{Opacity}, or  \textit{Location, Velocity}).
    \item \textbf{Relationships ($\mathcal{R}$):} The causal logical constraints between entities, formalized as \textit{Association}, \textit{Irrelevance}, and \textit{Exclusion}.
\end{itemize}

Taking the chest radiograph domain as a representative instantiation, as shown in Fig.~\ref{fig:know-graph}, the MLLM extracts structured knowledge from the MIMIC-CXR dataset~\cite{johnson2019mimic}. By leveraging iterative semantic parsing, 12 distinct disease entities are identified, accompanied by 53 clinically relevant attributes classified into chains of morphological, density, and anatomical categories. Fundamentally, these relationships serve as hard constraints for the subsequent generation of counterfactuals. This structured graph $\mathcal{G}_{K}$ defines the logical constraints for the reasoning space of the model.

\subsubsection{Counterfactual Thinking Trajectories}
\label{sec.3.5.2}

Instead of relying on an external large model, which incurs high computational latency and domain transfer costs, we propose a self-evolution strategy. Here, the target MLLM $\pi_{\theta}$ itself acts as the generator. 
Given an input image $v$ and a prompt $l$, the model first generates an initial thinking trajectory $t_{init}$. Subsequently, guided by the concept graph $\mathcal{G}_{K}$, the model performs controlled feature perturbation: it identifies key attributes in $t_{init}$ and substitutes them with causally distinct attributes defined in $\mathcal{G}_{K}$ to synthesize a set of candidate counterfactual reasoning paths $\mathcal{T}_{cand} = \{t^{'}_1, t^{'}_2, ..., t^{'}_N\}$. Through systematic integration of the concept graph, the MLLM $\pi$ itself evolves as an expert with causal prior knowledge of the specialized domain, and synthesizes counterfactual narratives through controlled feature perturbation while maintaining plausibility of physical reality.

\subsubsection{Counterfactual Perception Samples}
\label{sec.3.5.3}

Beyond counterfactual thinking trajectories, the causal decoupling in MLLMs fundamentally extends to the perception perspective, manifesting as the severe attention shift illustrated in Fig.~\ref{fig:visual-drift}. 
To resolve this multi-modal misalignment, the counterfactual causal disentanglement formulated in Section~\ref{sec:3.4} is also applied symmetrically to the visual inputs. To achieve this, we introduce the visual-reasoning consistency protocol designed to synthesize perception counterfactuals, thereby isolating the true causal effect of perception evidence from spurious correlations.



Specifically, leveraging the hierarchical domain knowledge graph $\mathcal{G}_{K}$, we first decompose the thinking trajectory of each training instance into discrete semantic attributes. Based on these attributes from original visual inputs $v_{init}$, we retrieve a set of visual counterfactual candidates $\mathcal{V}_{cf} = \{v^{'}_{1}, v^{'}_{2}, ..., v^{'}_{N}\}$. These candidates are selected via nearest-neighbor sampling in the attribute space, ensuring they share high-level semantic proximity with the original visual inputs while exhibiting distinct fine-grained visual features. This constructs a hard visual context where subtle details determine the ground truth.


To strictly filter for hard negatives, samples that induce genuine visual drifts, we employ the target MLLM $\pi_{\theta}$ itself as a consistency discriminator. Analogous to the self-counterfact reasoning in Section~\ref{sec.3.5.2}, we formulate an inverse matching task: given the ground-truth reasoning chain $t_{init}$, the model is tasked with identifying the correct corresponding image from the candidate pool $\{v_{init}\} \cup \mathcal{V}_{cf}$. If the MLLM fails to align the reasoning $t_{init}$ with the original image $v_{init}$ and instead assigns a higher matching probability to a distractor $v^{'}_{k} \in \mathcal{V}_{cf}$, it indicates a critical internal perception drift. Consequently, such mismatched pairs $(v^{'}_{k}, t_{init})$ are identified as hard counterfactual samples, which are counterfact perception shown in Fig.~\ref{fig:workflow}. These samples, possessing highly deceptive visual semantics, are explicitly integrated into the optimization objective to penalize endogenous perception drift.


\subsection{CPO++ for Multi-modal Disentanglement}
\label{sec:3.6}


Building upon the counterfactual causal disentanglement formulated in Section~\ref{sec:3.4} and the consistency protocol in Section~\ref{sec:3.5}, we propose CPO++ to drive the reinforced fine-tuning of the MLLM.

Formally, we have decomposed the generation process of CoT into a stream of next token prediction actions $S_{0,i} = \{s_{0},...,s_{i}\}$. Each cognitive state $s_{j}=(t_{<j},z_{j}|v)$ encapsulates the token sequence generated thus far $t_{<j}$ and the latent predicted distribution $z_{j}$ of the final results, strictly conditioned on the perception evidence $v$, as exhibited in Definition~\ref{def:multimodal_drift}. At timestep $j$, the action $t_{j}$ is sampled from the policy $\pi(\cdot | v, l, t_{<j})$ where $t_{j}$ can be any token in the vocabulary. After each action, the resulting state $s_{j+1}$ is the concatenation of the current state $s_{j}$ and the action $t_{j}$ with the corresponding latent predicted results:
\begin{equation}
s_{j+1}=(t_{<j} \circ t_{j}, P_{j}(z | t_{<j} \circ t_{j}, v)), 0\leq j \leq L,
\end{equation}
where $\circ$ denotes the concatenation between the token stream $t_{<j}$ and the action token $t_{j}$, $L$ represents the maximum length of the CoT, and $P_{j}$ is the latent predicted distribution of results derived by $t_{<j}$ based on the visual input $v$ as presented in Eq.~\eqref{eq:dis}. At the start of the thinking process, the initial action $t_{0}$ is typically the special token $<$think$>$. When the model produces the token $<$/think$>$, the resulting state $s_{L+1}$ becomes the terminal state, thereby concluding the generation process of one reasoning trajectory.

We regard the thinking trajectory that accurately grounds the perception evidence, such as the diagnostic report issued by professional doctors, as the positive sample preferred by humans. For rejected samples, we utilize the counterfactual trajectories, including the textual thinking sequences obtained in Section~\ref{sec.3.5.2}, as well as the hard negative perception samples in Section~\ref{sec.3.5.3}. These negative samples explicitly represent the endogenous reasoning drift, incorporating both thinking and perception perspectives. Thereby, the reasoning trajectory preferred by humans is defined as $t^{+}$, while the counterfactual trajectory exhibiting semantic or visual decoupling is represented by $ t^{-} \in \mathcal{T}_{cand} \cup \{ t_{init} \mid v \in \mathcal{V}_{cf} \}$.

Consequently, following the direct preference optimization approach~\cite{rafailovDirectPreferenceOptimization2023}, we derive the optimal policy that maximizes the reward function through:
\begin{equation}\pi_{\theta}(t|v,l) \propto \pi_{\text{ref}}(t|v,l)\exp\left(\frac{r(v,l,t)}{\beta}\right),
\end{equation}
where $\beta$ is a parameter controlling the deviation from the base reference policy $\pi_{\text{ref}}$, namely the initial supervised fine-tuning model, and $\pi_{\theta}$ denotes the model undergoing custom-tuning. Incorporating the counterfactual effect defined in Eq.~\eqref{Eq.4}, the difference in reward between the positive samples and the counterfactual samples can be defined as:
\begin{equation}
\begin{aligned}& r(v,l,t^{+}) - r(v,l,t^{-}) \ = \\ & \beta \left[ \log\frac{\pi_{\theta}(t^{+}|v,l)}{\pi_{\text{ref}}(t^{+}|v,l)} - \log\frac{\pi_{\theta}(t^{-}|v,l)}{\pi_{\text{ref}}(t^{-}|v,l)} \right]
\end{aligned}
\end{equation}

Thus, based on the Bradley-Terry model, the counterfactual preference optimization drives the reinforced custom-tuning of the MLLM through the following maximum likelihood objective:
\begin{equation}
\begin{aligned}
\mathcal{L}_{\text{CPO}} (\pi{\theta};\pi_{\text{ref}}) &= - \mathbb{E}{(v,l,t^{+},t^{-})} \Bigg[ \log \sigma \Bigg( \beta \log\frac{\pi{\theta}(t^{+}|v,l)}{\pi_{\text{ref}}(t^{+}|v,l)} \\
&\quad - \beta\log\frac{\pi_{\theta}(t^{-}|v,l)}{\pi_{\text{ref}}(t^{-}|v,l)} \Bigg) \Bigg]
\end{aligned}
\end{equation}

Ultimately, this formulation culminates in counterfactual reinforced custom-tuning. It serves as an adaptive framework that effectively differentiates between advantageous domain adaptation and detrimental spurious correlations. By integrating causal intervention with dynamic policy reinforcement, the proposed method actively corrects multi-modal concept drift, across both thinking and perception perspectives, adapting the MLLMs with endogenous reasoning drift.

\begin{table}[htbp]
\centering
\caption{\textbf{Experimental settings of tasks, domain, datasets (Size) and metrics.}
The experiments are initiated with the reinforced fine-tuning and rigorously evaluated across four distinct analytical dimensions: reasoning, decision, robustness, and generalization, within two critical sectors: the medical domain (Med), and autonomous driving (AD). 
The ultimate performance is comprehensively assessed utilizing a combination of NLP metrics, top-1 accuracy, and advanced GPT-Score evaluations to ensure precise measurement of logical consistency.}
\label{tab:oveall}
\renewcommand\arraystretch{1.5}
\setlength{\tabcolsep}{2mm}{
\begin{tabular}{@{}ccccc@{}}
\toprule
Task                                                                                        & Domain                     & Dataset                         & Size                                                                                    & Metrics                                      \\ \midrule
                                                                                            & Med                        & MIMIC-CXR                       & \begin{tabular}[c]{@{}c@{}}368K\\      (Finetune)\end{tabular}                          &                                              \\
\multirow{-2}{*}{\begin{tabular}[c]{@{}c@{}}Reinforced\\      Fine-tuning\end{tabular}}     & \cellcolor[HTML]{F2F2F2}AD & \cellcolor[HTML]{F2F2F2}BDD-X   & \cellcolor[HTML]{F2F2F2}\begin{tabular}[c]{@{}c@{}}5,597\\      (Finetune)\end{tabular} & \cellcolor[HTML]{F2F2F2}                     \\ \midrule
                                                                                            & Med                        & MIMIC-CXR                       & 5,159                                                                                   &                                              \\
\multirow{-2}{*}{Reasoning}                                                                 & \cellcolor[HTML]{F2F2F2}AD & \cellcolor[HTML]{F2F2F2}BDD-X   & \cellcolor[HTML]{F2F2F2}656                                                             & \cellcolor[HTML]{F2F2F2}\multirow{-2}{*}{NLP Metrics} \\ \midrule
                                                                                            & Med                        & MS-CXR-T                        & 1,326                                                                                   & Top-1 Acc.                                   \\
\multirow{-2}{*}{Decision}                                                                  & \cellcolor[HTML]{F2F2F2}AD & \cellcolor[HTML]{F2F2F2}BDD-X   & \cellcolor[HTML]{F2F2F2}656                                                             & \cellcolor[HTML]{F2F2F2}NLP Metrics          \\ \midrule
& Med                        & MS-CXR-T                                &   1,326                                                                                      &  Top-1 Acc.                   \\
\multirow{-2}{*}{Robustness}                                                                & \cellcolor[HTML]{F2F2F2}AD & \cellcolor[HTML]{F2F2F2}CODA-LM & \cellcolor[HTML]{F2F2F2}500                                                             & \cellcolor[HTML]{F2F2F2}GPT-Score            \\ \bottomrule
&                            & Open-I                          & 1,496                                                                                   &                                              \\
                                                                                            &                            & PadChest                        & 48K                                                                                     &                                              \\
                                                                                            &                            & ChestXray14                     & 25K                                                                                     &                                              \\
                                                                                            & \multirow{-4}{*}{Med}      & ChestXDet10                     & 661                                                                                     & \multirow{-4}{*}{Top-1 Acc.}                 \\
\multirow{-5}{*}{\begin{tabular}[c]{@{}c@{}}Generalization\\      (Zero-Shot)\end{tabular}} & \cellcolor[HTML]{F2F2F2}AD & \cellcolor[HTML]{F2F2F2}DriveLM & \cellcolor[HTML]{F2F2F2}216                                                             & \cellcolor[HTML]{F2F2F2}GPT-Score            \\ \bottomrule

\end{tabular}
}
\end{table}

\begin{table*}[htbp]
\centering
\caption{\textbf{Quantitative evaluation of diagnostic reasoning on the medical MIMIC-CXR~\cite{johnson2019mimic}  dataset.} It reports comprehensive metrics, including the BLEU-1/-2/-3/-4, ROUGE-L, METEOR and CIDEr on diagnostic report generation task, to assess the quality and coherence of the generated chain-of-thought trajectories. Red and blue highlight the best and second-best results, respectively.}
\label{tab:mimic-reason}
\begin{tabular}{@{}lcccccccc@{}}
\toprule
Methods                                                       & Venue      & BLEU-1                       & BLEU-2                       & BLEU-3                       & BLEU-4                       & ROUGE-L                      & METEOR                       & CIDEr                        \\ \midrule
\multicolumn{9}{l}{\textit{\textbf{Traditional}}}                                                                                                                                                                                                                                                   \\
GSK \cite{yang2022knowledge}                                  & MIA'22     & 0.363                        & 0.228                        & 0.156                        & 0.115                        & 0.284                        & -                            & 0.203                        \\
Clinical-BERT   \cite{yan2022clinical}                        & AAAI'22    & 0.383                        & 0.230                        & 0.151                        & 0.106                        & 0.275                        & 0.144                        & 0.151                        \\
METransformer   \cite{wang2023metransformer}                  & CVPR'23    & 0.386                        & 0.250                        & 0.169                        & 0.124                        & 0.291                        & 0.152                        & 0.362                        \\
STREAM \cite{yang2025spatio}                                  & TMI'25     & 0.420                        & 0.267                        & 0.184                        & 0.133                        & 0.291                        & 0.164                        &  -                            \\
DC-SGC   \cite{wangDiagnosticCaptioningCooperative2025}       & TPAMI'25   & 0.387                        & 0.251                        & 0.177                        & 0.129                        & 0.291                        & 0.164                        & 0.342                        \\ \midrule
\multicolumn{9}{l}{\textit{\textbf{LLM-Based}}}                                                                                                                                                                                                                                                     \\
R2GenGPT   \cite{wang2023r2gengpt}                            & MetaRad'23 & 0.408                        & 0.256                        & 0.174                        & 0.125                        & 0.285                        & 0.167                        & 0.244                        \\
PromptMRG   \cite{jin2024promptmrg}                           & AAAI'24    & 0.398                        & -                            & -                            & 0.112                        & 0.268                        & 0.157                        & -                            \\
BtspLLM   \cite{liu2024bootstrapping}                         & AAAI'24    & 0.402                        & 0.262                        & 0.180                        & 0.128                        & 0.291                        & 0.175                        & -                            \\
MambaXray   \cite{wangCXPMRGBenchPretrainingBenchmarking2024} & CVPR‘25    & 0.422                        & 0.268                        & 0.184                        & 0.133                        & 0.289                        & 0.167                        & 0.241                        \\ \midrule
\multicolumn{9}{l}{\textit{\textbf{Reinforced Reasoning}}}                                                                                                                                                                                                                                          \\
DPO\cite{rafailovDirectPreferenceOptimization2023}            & NeurIPS'23 & 0.409                        & 0.256                        & 0.183                        & 0.154                        & 0.316                        & {\color[HTML]{4472C4} 0.260} & 0.352                        \\
BoxMed-RL   \cite{jingReasonRadiologistChainofthought2026}    & MIA'25     & 0.426                        & 0.271                        & 0.185                        & 0.134                        & 0.314                        & 0.180                        & -                            \\
MPO \cite{xiao2025radiology}                                  & AAAI'25    & 0.416                        & 0.269                        & {\color[HTML]{4472C4} 0.191} & 0.139                        & 0.309                        & 0.162                        & -                            \\
FiR-Rad \cite{mei2026fir}                                     & TMI'26     & -                            & -                            & -                            & {\color[HTML]{4472C4} 0.158} & {\color[HTML]{4472C4} 0.326} & 0.181                        & -                            \\
CPO \cite{yang2025walking}                                    & NeurIPS'25 & {\color[HTML]{4472C4} 0.426} & {\color[HTML]{FF0000} 0.288} & 0.186                        & 0.155                        & 0.321                        & 0.236                        & {\color[HTML]{4472C4} 0.375} \\
\rowcolor[HTML]{D9D9D9} 
CPO++                                                         &            & {\color[HTML]{FF0000} 0.429} & {\color[HTML]{4472C4} 0.286} & {\color[HTML]{FF0000} 0.193} & {\color[HTML]{FF0000} 0.165} & {\color[HTML]{FF0000} 0.329} & {\color[HTML]{FF0000} 0.271} & {\color[HTML]{FF0000} 0.378} \\ \bottomrule
\end{tabular}
\end{table*}

\section{Experiments}


This section presents a comprehensive evaluation of the proposed framework. First, Section \ref{sec:exp_set} details the experimental setup and implementation details. Section \ref{sec:exp_reason} investigates the capability to perform reasoning under non-stationary conditions. Building upon this, Section \ref{sec:exp_decision} introduces the reliable decision-making derived from the reasoning. Furthermore, Section \ref{sec:exp_Robust} validates the endogenous robustness of the proposed method during reasoning and decision-making. Subsequently, the Section \ref{sec:exp_generalization} evaluates the generalization performance extrapolated from this endogenous robustness. Finally, Section \ref{sec:exp_abl} provides ablation studies to demonstrate the critical role of visual and textual components in maintaining the internal robustness.


\subsection{Experiments Settings}
\label{sec:exp_set}

To comprehensively evaluate the proposed framework, experiments are conducted across two distinct sectors: medical chest X-ray diagnosis and autonomous driving. These domains are strategically selected because they represent real-world safety-critical environments that demand rigorous vision-language reasoning and strict logical consistency. Furthermore, both fields frequently exhibit highly non-stationary conditions, thereby providing an optimal testbed to validate the effectiveness of the proposed approach. A detailed summary of the datasets and evaluation metrics utilized across fine-tuning phases and all downstream tasks is provided in Table~\ref{tab:oveall}.

Regarding the implementation details, Qwen2.5-VL (7B)~\cite{qwen2.5-VL} is adopted as the default backbone architecture unless otherwise specified. The models are fine-tuned for one epoch with the batch size of 4. All training procedures are executed on a hardware setup comprising $2\times2$ NVIDIA A100 GPUs. The optimization process is driven by the AdamW optimizer, configured with an initial learning rate of 2e-5 and a warm-up period of 40 steps to ensure stable convergence.

\subsection{Taming the Non-stationary Reasoning}
\label{sec:exp_reason}

This section rigorously evaluates the multi-modal reasoning capabilities of the proposed framework across both the medical and autonomous driving domains.

In the medical domain, diagnostic reports serve as the gold standard for evaluating the reasoning process of various models. The quantitative results are detailed in Table \ref{tab:mimic-reason}. 
First, it reveals that while LLM-based methods achieve improvements in fundamental metrics such as BLEU-1 and BLEU-2 over traditional approaches, the performance of these models remains limited on more complex structural metrics. This limitation indicates that relying solely on the language prior of LLMs is insufficient, thereby underscoring the necessity of reinforcement fine-tuning to enhance endogenous reasoning robustness. 
Furthermore, the proposed CPO++ demonstrates a substantial advantage in higher-order semantic metrics, achieving scores of 0.271 in METEOR and 0.378 in CIDEr. These results confirm that the framework conducts causally-driven reasoning consistent with the logic of medical professionals, rather than simply memorizing clinical terminology. Although the result on BLEU-2 marginally trails behind one baseline, the proposed approach exhibits significantly stronger logical coherence when compared to other reinforced reasoning methods. This overall superiority explicitly highlights the critical importance of integrating counterfactual causality as detailed in Section \ref{sec:3.6}.

\begin{figure}[htbp]
    \centering
    \includegraphics[width=1\linewidth]{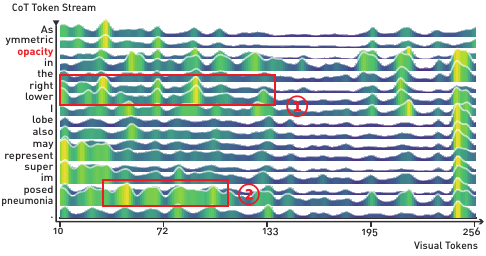}
    \caption{\textbf{Qualitative evaluation of the attention scores over visual tokens during CoT decoding.} When the term ’lung opacity’ is subtly altered to ’opacity’, the model still produces high responses in key areas, such as the visual tokens at the right side of \ding{172} and the pneumonia \ding{173}. }
    \label{fig:case-study}
\end{figure}

To qualitatively evaluate the effectiveness of the proposed framework in mitigating endogenous reasoning drift, a visualization of cross-modal attention scores is presented in Fig.~\ref{fig:case-study}. This case mirrors the critical instability previously observed in Fig.~\ref{fig:visual-drift}, wherein the subtle substitution of 'lung opacity' with 'opacity' triggered a systemic collapse of visual grounding in the baseline model. In stark contrast, even when subjected to identical token perturbations, the policy optimized through CPO++ maintains robust attention responses across critical pathological regions. As highlighted by markers \ding{172} and \ding{173} in Fig.~\ref{fig:case-study}, the model remains strictly anchored to visual tokens representing the lower right lobe and the manifestation of pneumonia. This result confirms that the counterfactual decoupling mechanism effectively stabilizes the perception perspective, ensuring that the cognitive process of the model remains anchored in objective visual evidence rather than being hijacked by linguistic biases.

\begin{table*}[htbp]
\centering
\caption{\textbf{Quantitative evaluation of driving reasoning on the autonomous driving BDD-X \cite{kim2018textual} dataset.} It reports metrics, including BLEU-4, ROUGE-L, and CIDEr, to assess the logical correctness and linguistic quality of the generated reasoning trajectories. Following the evaluation protocol of DriveGPT4 \cite{xu2024drivegpt4}, the driving scenarios are systematically categorized into Easy, Medium, and Hard splits to evaluate the reasoning capabilities under varying degrees of environmental complexity. Red and blue indicate the best and second-best results, respectively.}
\label{tab:reason-ad}
\setlength{\tabcolsep}{1.3mm}{
\begin{tabular}{@{}lccccccccccccc@{}}
\toprule
                                   &                         & \multicolumn{3}{c}{Easy}                                                                   & \multicolumn{3}{c}{Medium}                                                                 & \multicolumn{3}{c}{Hard}                                                                   & \multicolumn{3}{c}{All}                                                                    \\
\multirow{-2}{*}{Methods}          & \multirow{-2}{*}{Venue} & BLEU-4                       & ROUGE-L                      & CIDEr                        & BLEU-4                       & ROUGE-L                      & CIDEr                        & BLEU-4                       & ROUGE-L                      & CIDEr                        & BLEU-4                       & ROUGE-L                      & CIDEr                        \\ \midrule
ADAPT \cite{jin2023adapt}          & ICRA'23                 & {\color[HTML]{4472C4} 0.209} & 0.462                        & 1.009                        & 0.164                        & 0.408                        & 0.627                        & {\color[HTML]{4472C4} 0.136} & 0.405                        & 0.527                        & 0.174                        & 0.430                        & 0.854                        \\
LLaVA-1.5   \cite{liu2024improved} & CVPR'24                 & 0.198                        & 0.476                        & 1.118                        & 0.147                        & 0.403                        & 0.638                        & 0.116                        & 0.420                        & 0.498                        & 0.174                        & 0.454                        & 0.965                        \\
DriveGPT4   \cite{xu2024drivegpt4} & RAL'24                  & 0.204                        & 0.465                        & 1.132                        & 0.169                        & 0.405                        & 0.650                        & 0.123                        & 0.421                        & 0.573                        & {\color[HTML]{4472C4} 0.183} & 0.447                        & 0.991                        \\
HoP \cite{zhou2025hints}           & ICCV'25                 & 0.201                        & 0.476                        & {\color[HTML]{FF0000} 1.210} & {\color[HTML]{4472C4} 0.180} & 0.408                        & 0.652                        & 0.130                        & {\color[HTML]{4472C4} 0.435} & 0.604                        & 0.179                        & {\color[HTML]{000000} 0.458} & {\color[HTML]{4472C4} 1.022} \\
CPO \cite{yang2025walking}         & NeurIPS'25              & 0.207                        & {\color[HTML]{4472C4} 0.480} & 1.180                        & 0.176                        & {\color[HTML]{4472C4} 0.422} & {\color[HTML]{4472C4} 0.667} & 0.135                        & 0.429                        & {\color[HTML]{4472C4} 0.624} & 0.182                        & {\color[HTML]{4472C4} 0.462} & 0.990                        \\
\rowcolor[HTML]{D9D9D9} 
CPO++                              &                         & {\color[HTML]{FF0000} 0.211} & {\color[HTML]{FF0000} 0.491} & {\color[HTML]{4472C4} 1.208} & {\color[HTML]{FF0000} 0.190} & {\color[HTML]{FF0000} 0.441} & {\color[HTML]{FF0000} 0.689} & {\color[HTML]{FF0000} 0.156} & {\color[HTML]{FF0000} 0.458} & {\color[HTML]{FF0000} 0.642} & {\color[HTML]{FF0000} 0.195} & {\color[HTML]{FF0000} 0.477} & {\color[HTML]{FF0000} 1.045} \\ \bottomrule
\end{tabular}
}
\end{table*}

Within the autonomous driving domain, the evaluation follows the protocol established by DriveGPT4 by systematically categorizing the reasoning scenarios into Easy, Medium, and Hard subsets, as presented in Table \ref{tab:reason-ad}. As the environmental complexity transitions from the Easy split to the Hard split, baseline methods experience a severe performance degradation. For instance, the CIDEr score of ADAPT plummets from 1.009 down to 0.527. In stark contrast, the proposed CPO++ exhibits exceptional resilience under the Hard setting by maintaining a high CIDEr score of 0.642, which directly validates the capacity of the model to handle non-stationary dynamics. Consistent with the analysis in the medical domain, the strong performance in the CIDEr metric emphasizes that the model performs logical deductions akin to a human driver, rather than generating superficial visual descriptions. While the CIDEr score of the proposed method in the Easy subset (1.208) is slightly lower than that of the HoP baseline (1.210), the overwhelming superiority across the Hard subset and the overall average definitively confirms the advantage of the approach in processing complex driving semantics.

\subsection{Anchoring the Reliable Decision-Making}
\label{sec:exp_decision}

Building upon the logical reasoning capabilities established in the above Section \ref{sec:exp_reason}, the evaluation now transitions to assessing the capacity of the framework to formulate reliable downstream decisions.

\begin{table}[htbp]
\centering
\caption{\textbf{Quantitative evaluation of diagnostic decision-making on the medical MS-CXR-T\cite{bannurLearningExploitTemporal2023} dataset.} It reports the Top-1 accuracy on mutli-label classification task across five core pulmonary pathologies, including consolidation (Con.), pleural effusion (PE), pneumonia (Pna.), pneumothorax (Pnx.), and Edema (Ede.). Red and blue highlight the best and second-best results, respectively.}
\label{tab:class-medical}
\setlength{\tabcolsep}{1.6mm}{
\begin{tabular}{@{}lccccccc@{}}
\toprule
                                                            & Venue      & Con.                        & PE                          & Pna.                        & Pnx.                        & Ede.                        & Avg.                        \\ \midrule
CheXRelNet  \cite{karwande2022chexrelnet}                   & MICCAI'22  & 47.0                        & 47.0                        & 47.0                        & 36.0                        & 49.0                        & 45.2                        \\
BioViL \cite{boecking2022making}                            & ECCV'22    & 56.0                        & 63.0                        & 60.2                        & 42.5                        & 67.5                        & 57.8                        \\
CTrans \cite{Bannur_2023_CVPR}                        & CVPR'23    & 44.0                        & 61.3                        & 45.1                        & 31.5                        & 65.5                        & 49.5                        \\
BioViL-T    \cite{Bannur_2023_CVPR}                         & CVPR'23    & 61.1                        & 67.0                        & 61.9                        & 42.6                        & 68.5                        & 60.2                        \\
Med-ST \cite{yang2024unlocking}                             & ICML'24    & 60.6                        & 67.4                        & 58.5                        & 65.0                        & 54.2                        & 61.1                        \\
TempA-VLP \cite{10943948}                                   & WACV'25    & 65.2                        & 59.4                        & 73.4                        & 43.1                        & {\color[HTML]{4472C4} 77.1} & 63.6                        \\
CoCa-CXR   \cite{chen2025cocacxrcontrastivecaptionerslearn} & MICCAI'25  & 70.4                        & 69.6                        & 61.4                        & 72.8                        & 71.8                        & 69.2                        \\ \midrule
SFT                                                         &            & 54.9                        & 71.7                        & 70.0                        & {\color[HTML]{FF0000} 95.9} & 76.5                        & 73.8                        \\
DPO   \cite{rafailovDirectPreferenceOptimization2023}       & NeurIPS'23 & 63.2                        & 72.4                        & 76.7                        & 93.5                        & 76.3                        & 76.4                        \\
CPO \cite{yang2025walking}                                  & NeurIPS'25 & {\color[HTML]{FF0000} 77.7} & {\color[HTML]{4472C4} 72.7} & {\color[HTML]{4472C4} 87.4} & {\color[HTML]{4472C4} 95.8} & 75.3                        & {\color[HTML]{4472C4} 81.8} \\
\rowcolor[HTML]{D9D9D9} 
CPO++                                                       &            & {\color[HTML]{4472C4} 77.0} & {\color[HTML]{FF0000} 78.1} & {\color[HTML]{FF0000} 87.6} & 95.4                        & {\color[HTML]{FF0000} 78.4} & {\color[HTML]{FF0000} 83.3} \\ \bottomrule
\end{tabular}
}
\end{table}

Table~\ref{tab:class-medical} presents the quantitative results of the diagnostic decision-making task on the medical MS-CXR-T dataset. An overarching analysis reveals that the proposed CPO++ achieves the highest average accuracy, substantially outperforming the majority of the evaluated methods.
A detailed comparison against baseline models further highlights the specific advantages of the proposed architecture. First, current LLM-based models demonstrate a fundamental performance ceiling, with average scores consistently failing to reach 70.0. Subsequently, while the standard DPO approach yields a higher accuracy than the SFT baseline, the enhancement is extremely limited, increasing from 73.8 to merely 76.4. This marginal gain directly proves that standard DPO remains severely constrained by language priors and exhibits a high vulnerability to multi-modal concept drift during complex medical classification tasks, corroborating the preliminary observations presented in Figure 1. 
Moreover, a direct comparison between the proposed approach and the prior CPO method indicates a clear performance leap. This improvement conclusively demonstrates that the counterfactual decoupling of both visual and textual modalities, as discussed in Section~\ref{sec:3.5}, is fundamentally more effective than the isolated decoupling of language reasoning.

Within the autonomous driving domain, the evaluation underscores the severe complexity of making decisions in highly dynamic scenarios. The experimental results on the BDD-X dataset are summarized in Table~\ref{tab:decision-ad}, leveraging NLP metrics to rigorously measure the precision of the generated complex driving actions. 
The results indicate that the proposed framework achieves dominant performance across the BLEU-4 and METEOR metrics. This comprehensive superiority confirms the exceptional reliability of the framework in synthesizing accurate and executable driving instructions, even when confronted with complex and highly dynamic environmental changes.

\begin{table}[htbp]
\centering
\caption{\textbf{Quantitative evaluation of driving decision-making on the autonomous driving BDD-X~\cite{kim2018textual} dataset.} It reports metrics, including BLEU-4, METEOR, and CIDEr, to evaluate the accuracy of the generated driving actions (\eg, "The car merges into the lane to its left"). Red and blue indicate the best and second-best results, respectively.}
\label{tab:decision-ad}
\begin{tabular}{@{}lcccc@{}}
\toprule
Methods                               & Venue   & BLEU-4                       & METEOR                       & CIDEr                        \\ \midrule
WAA \cite{kim2018textual}             & ECCV'18 & 0.323                        & 0.292                        & 2.158                        \\
ADAPT \cite{jin2023adapt}             & ICRA'23 & 0.348                        & 0.305                        & 2.499                        \\
DriveGPT4   \cite{xu2024drivegpt4}    & RAL'24  & 0.300                        & 0.298                        & 2.140                        \\
RAG-Driver \cite{yuan2024rag}         & RSS'24  & 0.343                        & {\color[HTML]{4472C4} 0.307} & 2.608 \\
Dolphins \cite{ma2024dolphins}        & ECCV'24 & 0.306                        & 0.282                        & 2.236                        \\
MCAM \cite{cheng2025mcam}             & ICCV'25 & {\color[HTML]{4472C4} 0.357} & 0.305                        & 2.520                        \\
CPO \cite{yang2025walking} & NeurIPS'25 & 0.355 & 0.298 & {\color[HTML]{4472C4} 2.631} \\
\rowcolor[HTML]{D9D9D9} 
CPO++                                 &         & {\color[HTML]{FF0000} 0.363} & {\color[HTML]{FF0000} 0.319} & {\color[HTML]{FF0000} 2.734} \\ \bottomrule
\end{tabular}
\end{table}

\subsection{Fortifying the Endogenous Robustness}
\label{sec:exp_Robust}

Having established the efficacy of the framework in logical reasoning and decision-making, as detailed in the above sections, the subsequent evaluation shifts focus to the endogenous robustness of the model under complex and non-stationary conditions.

\begin{figure}[htbp]
    \centering
    \includegraphics[width=0.48\textwidth]{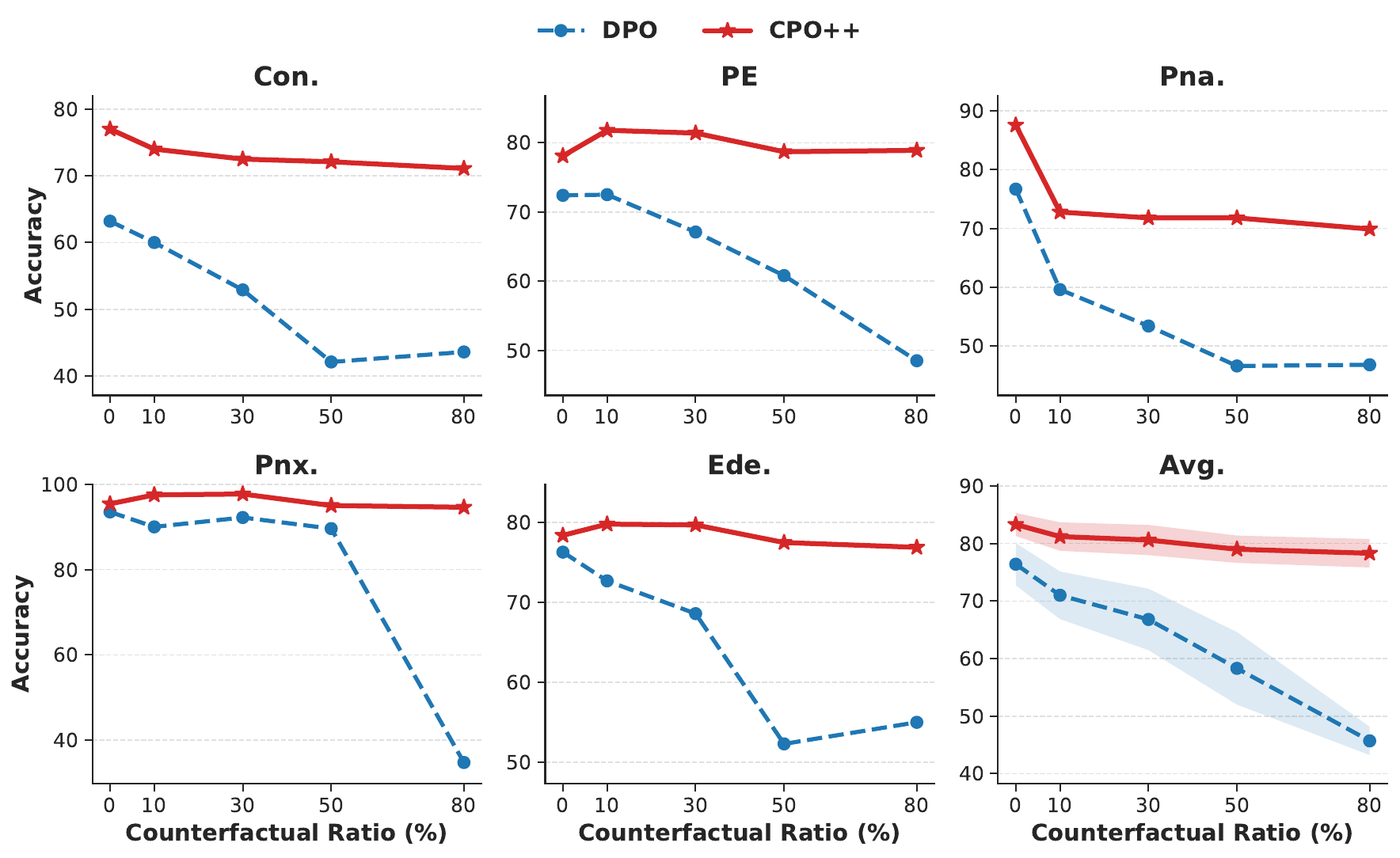}
    \caption{\textbf{Quantitative evaluation of diagnostic robustness against counterfactual interference on the medical MS-CXR-T \cite{bannurLearningExploitTemporal2023} dataset.} It reports the Top-1 accuracy across five pulmonary pathologies, including consolidation (Con.), pleural effusion (PE), pneumonia (Pna.), pneumothorax (Pnx.), and Edema (Ede.) and their overall average (Avg.).To simulate non-stationary and complex reasoning, varying ratios of counterfactual reasoning interference are injected into the prompts during the inference phase. }
    \label{fig:exp-med-rob}
\end{figure}

First, we conduct experiments in the medical diagnosis scenario, that deliberately disrupt the reasoning process by injecting varying ratios of counterfactual tokens into the prompts. The results are exhibited in Fig. \ref{fig:exp-med-rob}. The average performance reveals the severe vulnerability of the current reinforced reasoning model, such as DPO. As the ratio of counterfact increases, the performance of the DPO baseline experiences a catastrophic decline. 
This phenomenon demonstrates that reinforced models are easily hijacked by cumulative drift within the reasoning trajectory, as illustrated in Fig. \ref{fig:toy-concept}. 
Specifically, for pathologies with highly distinct visual features, such as pneumothorax, DPO maintains stability under minor interference but suffers a complete reasoning collapse under severe disruption with 80\% counterfactual perturbation ratio.

 In stark contrast, the proposed CPO++ framework demonstrates significant stability. The performance curve of the proposed approach remains nearly horizontal even under an extreme interference ratio of 80\%. This unwavering resilience proves that the framework genuinely grounds the reasoning process in visual evidence rather than blindly following misleading reasoning prompts.

\begin{table}[htbp]
\centering
\caption{\textbf{Quantitative evaluation of robustness on the autonomous driving CODA-LM \cite{chen2025automated} dataset.} It reports the GPT-Score across three comprehensive aspects: General Perception (General), Corner Case Perception (Corner), and Driving Suggestion (Suggest). Red and blue indicate the best and second-best results, respectively.}
\label{tab:exp-ad-rob}
\setlength{\tabcolsep}{1mm}{
\begin{tabular}{@{}lccccc@{}}
\toprule
Methods                                       & Venue                        & General & Corner & Suggest & Avg.                        \\ \midrule
GPT-4V \cite{openai2023gpt4}                  &                              & {\color[HTML]{FF0000} 57.5}                                       & 56.3                                                             & {\color[HTML]{4472C4} 63.3}                                       & 59.0                        \\
Gemini-Pro   \cite{comanici2025gemini}        &                              & 25.2                                                              & 51.4                                                             & 27.4                                                              & 34.7                        \\
Qwen-VL-MAX   \cite{yang2025qwen3}            & \multirow{-3}{*}{Commercial} & 34.6                                                              & 68.2                                                             & 47.4                                                              & 50.1                        \\ \midrule
InternVL   \cite{chen2024internvl}            & CVPR'24                      & 38.4                                                              & 61.5                                                             & 41.2                                                              & 47.0                        \\
LLaVA-NeXT   \cite{li2025llavanextinterleave} & ICLR'25                      & 29.9                                                              & 53.6                                                             & 31.9                                                              & 38.5                        \\
CODA-VLM   \cite{chen2025automated}           & WACV'25                      & 55.0                                                              & 77.7                                                             & 58.1                                                              & 63.6                        \\
Percept-DriveLM   \cite{sun2025towards}       & ICRA'25                      & 54.3                                                              & {\color[HTML]{4472C4} 83.1}                                                              & 60.5                                                              & {\color[HTML]{4472C4} 66.0} \\
MPDrive \cite{zhangMPDriveImprovingSpatial2025} & CVPR'25                      & 41.8                                                              & 76.1                                                             & 58.2                                                              & 58.7                        \\
CPO \cite{yang2025walking} & NeurIPS'25 & 48.2 & 82.3 & 60.9 & 63.5 \\ 
\rowcolor[HTML]{D9D9D9} 
CPO++                                         &                              & {\color[HTML]{4472C4} 55.6}                                       & {\color[HTML]{FF0000} 85.2}                                      & {\color[HTML]{FF0000} 65.5}                                       & {\color[HTML]{FF0000} 68.8} \\ \bottomrule
\end{tabular}
}
\end{table}

Furthermore, we conduct the evaluation that primarily focuses on robustness under extreme corner cases in autonomous driving, as shown in Table \ref{tab:exp-ad-rob}. 
Compared with commercial general-purpose models, while GPT-4V achieves the highest score in general perception, the performance of these commercial model exhibits significant weakness when confronting corner cases. 
Conversely, the proposed CPO++ framework achieves a remarkable and leading score of 85.2 in corner case perception. This substantial improvement is directly attributed to the implemented counterfactual intervention mechanism, which successfully resists visual drift when encountering long-tail and rare objects. Finally, the exceptional capability in extreme perception seamlessly translates into the highest score for driving suggestion, effectively completing the causal loop from perception to decision-making. Although the score of the proposed approach in general perception is marginally lower than that of GPT-4V, this objective outcome confirms that the framework perfectly balances robustness in extreme environments with general perceptual capabilities.


\subsection{Unleashing the Cross-Domain Generalization}
\label{sec:exp_generalization}

As demonstrated in the preceding evaluation, fortifying the inherent robustness of the architecture intrinsically cultivates a deeper understanding of underlying causal mechanisms. Driven by this correlation, the current section investigates the zero-shot generalization capabilities of the proposed framework. The primary objective is to prove that the robust representations acquired during training can seamlessly extrapolate to entirely unseen domains.

\begin{table}[htbp]
\centering
\caption{\textbf{Quantitative evaluation of zero-shot multi-label disease classification across diverse medical datasets.} The table reports the classification performance on four unseen external chest X-ray datasets: Open-I \cite{demner2012design}, PadChest \cite{bustos2020padchest}, ChestXray14 (CheX'14) \cite{Wang_2017_CVPR}, and CheXDet10 \cite{liu2020chestxdet10}. Red and blue indicate the best and second-best results, respectively.}
\label{tab:gen-med}
\setlength{\tabcolsep}{1.2mm}{
\begin{tabular}{@{}lccccc@{}}
\toprule
Method                                  & Venue         & Open-I                      & PadChest                    & CheX'14                 & CheXDet10                 \\ \midrule
CheXzero \cite{tiu2022expert}           & Nat Bio'22 & 75.9                        & 62.9                        & 72.6                        & 71.3                        \\
BioViL \cite{Bannur_2023_CVPR}          & CVPR'23       & 70.2                        & 65.5                        & 72.9                        & 70.8                        \\
MedKLIP \cite{wu2023medklip}            & ICCV'23       & 75.9                        & 62.9                        & 72.6                        & 71.3                        \\
KAD \cite{zhang2023knowledge}           & Nat Com'23 & 80.7                        & 75.0                        & 78.9                        & 73.5                        \\
CARZero \cite{lai2024carzero}           & CVPR'24       & 83.8                        & 81.0                        & 81.1                        & 79.6                        \\
CPO                                     & NeurIPS'25    & {\color[HTML]{4472C4} 84.4} & {\color[HTML]{4472C4} 82.0} & {\color[HTML]{4472C4} 81.7} & {\color[HTML]{4472C4} 80.1} \\
\rowcolor[HTML]{D9D9D9} 
CPO++                                   &               & {\color[HTML]{FF0000} 85.1} & {\color[HTML]{FF0000} 82.3} & {\color[HTML]{FF0000} 82.0} & {\color[HTML]{FF0000} 81.4} \\ \bottomrule
\end{tabular}
}
\end{table}

Within the medical diagnosis scenario, the zero-shot performance is rigorously assessed across multiple external datasets, with the quantitative results detailed in Table \ref{tab:gen-med}. Initially, the proposed CPO++ achieves the highest scores across all evaluated datasets. This widespread superiority indicates that the framework successfully extracts universal disease logic rather than merely overfitting to the specific data distribution of the MIMIC-CXR dataset. Furthermore, when confronted with highly heterogeneous external data, which inherently contains confounding biases originating from different hospitals, imaging equipment, and patient populations, the advantage of the upgraded CPO++ becomes evident. A direct comparison demonstrates that the CPO++ framework, which utilizes the counterfactual decoupling of both visual and textual modalities, exhibits a more robust generalization capability than the prior CPO method.

\begin{table}[htbp]
\centering
\caption{\textbf{Quantitative evaluation of zero-shot generalization on the autonomous driving DriveLM \cite{sima2024drivelm} dataset.} Following the evaluation protocol established by Dolphins, it reports zero-shot performance across three comprehensive aspects: Description (Desc.), Perception (Perc.), and Prediction \& Planning (Pred.\&Plan). Red and blue indicate the best and second-best results, respectively.}
\label{tab:gen-ad}
\begin{tabular}{@{}lccccc@{}}
\toprule
Models                                             & Venue    & Desc.                 & Perc.                  & Pred.\&Plan & Avg.                         \\ \midrule
Video-ChatGPT   \cite{maaz2024video}               & ACL'24   & 16.8                        & 34.8                        & 37.7                                                                  & 29.8                        \\
Dolphins   \cite{maDolphinsMultimodalLanguage2025} & ECCV'24  & {\color[HTML]{4472C4} 29.1} & {\color[HTML]{4472C4} 44.5} & {\color[HTML]{4472C4} 41.0}                                           & {\color[HTML]{4472C4} 38.2} \\
OpenFlamingo   \cite{gadetsky2025large}            & ICLR'25  & 6.9                         & 24.9                        & 22.4                                                                  & 18.1                        \\
Otter-Video \cite{li2025otter}                     & TPAMI'25 & 10.6                        & 37.8                        & 28.3                                                                  & 25.6                        \\
CPO \cite{yang2025walking} & NeurIPS'25 & 28.8 & 42.0 & 38.4 & 36.4 \\
\rowcolor[HTML]{D9D9D9} 
CPO++   &          & {\color[HTML]{FF0000} 30.0} & {\color[HTML]{FF0000} 46.2} & {\color[HTML]{FF0000} 43.8}                                           & {\color[HTML]{FF0000} 40.0} \\ \bottomrule
\end{tabular}
\end{table}

Similarly, in the autonomous driving scenario, the generalization capacity is comprehensively tested across three critical levels: description, perception, and prediction \& planning, as presented in Table \ref{tab:gen-ad}. An analysis of the baselines reveals that general-purpose video foundation models, such as OpenFlamingo and Video-ChatGPT, exhibit exceedingly poor performance in professional driving scenarios due to a fundamental lack of causal alignment. While dedicated autonomous driving models, represented by Dolphins, deliver acceptable results, the proposed CPO++ continues to establish a significant performance gap. This substantial margin confirms that the causally aligned representations acquired by the framework can be reliably generalized to interpret complex dynamics in unfamiliar traffic environments.


\begin{figure}[htbp]
    \centering
    \includegraphics[width=0.4\textwidth]{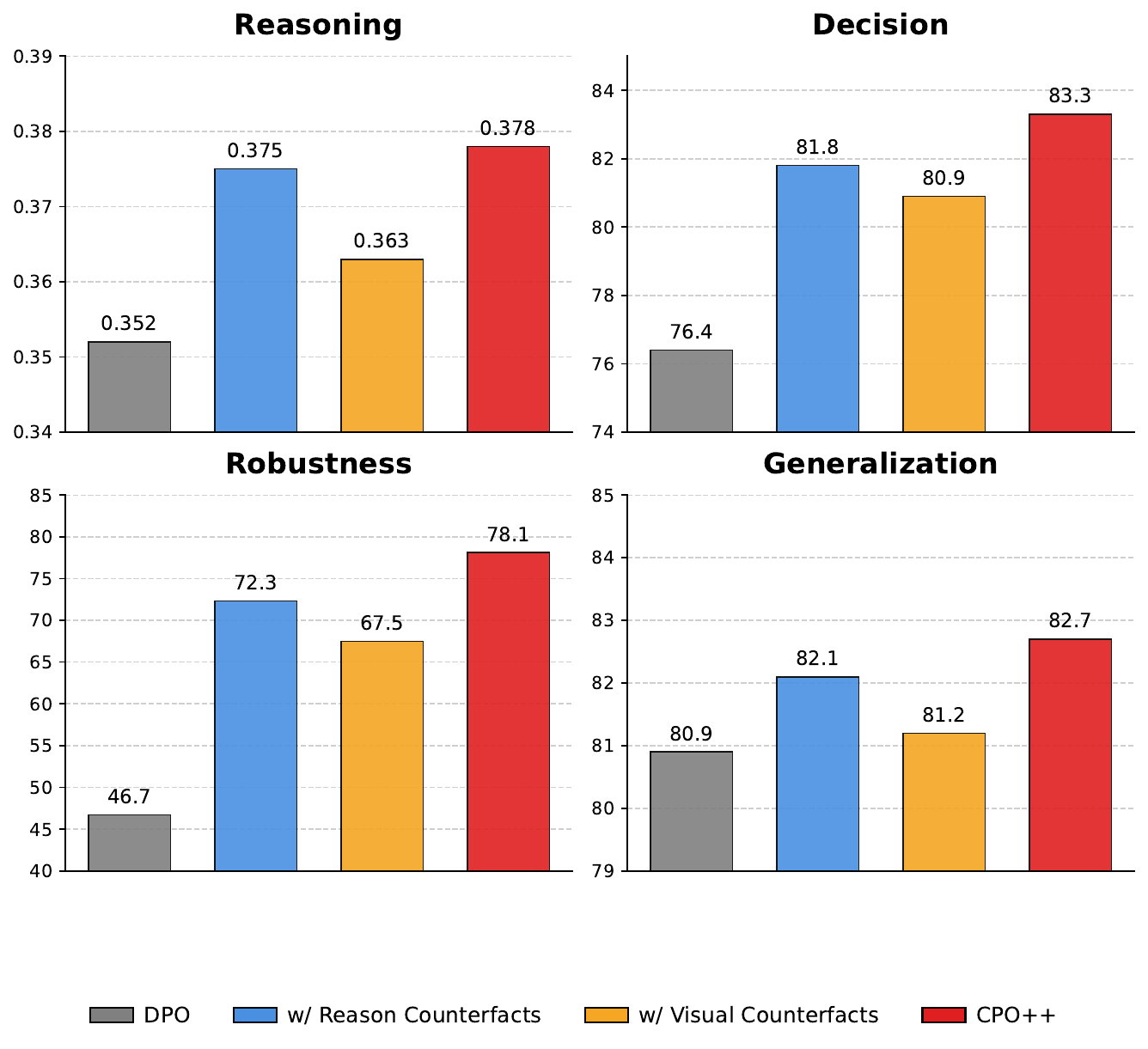}
    \caption{\textbf{Ablation study of the counterfactual decoupling mechanism.} The evaluation systematically compares the DPO baseline against isolated interventions, specifically integrating only reasoning counterfactuals or only visual counterfactuals, alongside the complete CPO++ framework featuring dual alignment. Performance is rigorously measured across four critical dimensions: 1) reasoning capability, evaluated via the CIDEr metric on the MIMIC-CXR dataset, 2) decision-making proficiency, represented by average accuracy on the MS-CXR-T dataset, 3) inherent robustness, validated by the average accuracy under an extreme 80\% counterfactual ratio on the MS-CXR-T dataset, and 4) cross-domain generalization, presented by average accuracy in zero-shot evaluations. }
    \label{fig:exp-abl}
\end{figure}

\subsection{Ablation Studies on Counterfactual Decoupling}
\label{sec:exp_abl}

To rigorously validate the individual contributions of the proposed visual and textual counterfactual decoupling mechanisms, a comprehensive ablation study is conducted. As illustrated in Fig.\ref{fig:exp-abl}, the evaluation is systematically measured across four critical dimensions: reasoning, decision-making, robustness, and generalization. 

In complex non-stationary environments, suppressing the reasoning drift and language priors of the LLM yields highly intuitive logical improvements. Simultaneously, the isolated visual decoupling mechanism independently contributes a substantial performance increase, definitively proving that both modules are indispensable. Furthermore, the experimental results demonstrate that the visual feature decoupling and the textual logic decoupling are not redundant. Instead, these two mechanisms operate in a strictly orthogonal and synergistic manner. Ultimately, only the simultaneous execution of this dual alignment process can push the overall performance of the framework to the theoretical upper limit.



\section{Conclusion}

In this work, we reveal a critical vulnerability within MLLMs known as endogenous reasoning drift, which is characterized by unpredictable distribution changes that emerge spontaneously during the cognitive process. To establish a rigorous foundation, we formalize this phenomenon as multi-modal concept drift. To mitigate this instability, we propose Counterfactual Preference Optimization ++ (CPO++), a comprehensive framework that integrates counterfactual reasoning with domain knowledge to execute controlled perturbations. By leveraging preference optimization, CPO++ effectively disentangles spurious correlations from genuine causal logic. Extensive empirical evaluations across two safety-critical domains, namely medical diagnosis and autonomous driving, demonstrate that the proposed framework achieves superior performance. In particular, CPO++ significantly enhances reasoning coherence and inherent robustness, providing a foundation for reliable multi-modal reasoning in high-stakes applications.

\section*{Acknowledgment}
The work was supported by the Australian Research Council (ARC) under Laureate project FL190100149.


\bibliographystyle{IEEEtran}
\bibliography{egbib}

\end{document}